\documentclass[11pt]{article}

\usepackage[preprint]{acl}

\usepackage{times}
\usepackage{latexsym}

\usepackage[T1]{fontenc}

\usepackage[utf8]{inputenc}

\usepackage{microtype}

\usepackage{inconsolata}

\usepackage{graphicx}

\usepackage{enumitem}
\usepackage{amsmath}
\usepackage{tabularx}
\usepackage{booktabs}
\usepackage{multirow}
\usepackage{colortbl}
\usepackage{makecell}
\usepackage{algorithmic}
\usepackage{amssymb}
\usepackage[colorinlistoftodos]{todonotes}

\newcommand{\best}[1]{\textbf{#1}}
\newcommand{\second}[1]{\underline{#1}}

\title{KnowledgeBerg: Evaluating Systematic Knowledge Coverage and Compositional Reasoning in Large Language Models}

\author{
Xiao Zhang \\
University of Groningen \\
\texttt{xiao.zhang@rug.nl}
\And
Qianru Meng \\
LIACS, Leiden University \\
\texttt{q.r.meng@liacs.leidenuniv.nl}
\And
Yongjian Chen \\
University of Groningen \\
\texttt{yongjian.chen@rug.nl}
\AND
Yumeng Wang \\
LIACS, Leiden University \\
\texttt{y.wang@liacs.leidenuniv.nl}
\And
Johan Bos \\
University of Groningen \\
\texttt{johan.bos@rug.nl}
}

\begin{document}
\maketitle
\begin{abstract}
Many real-world questions appear deceptively simple yet implicitly demand two capabilities: (i) systematic coverage of a bounded knowledge universe and (ii) compositional set-based reasoning over that universe, a phenomenon we term ``the tip of the iceberg.'' We formalize this challenge through two orthogonal dimensions: \emph{knowledge width}, the cardinality of the required universe, and \emph{reasoning depth}, the number of compositional set operations. We introduce \textsc{KnowledgeBerg}, a benchmark of 4,800 multiple-choice questions derived from 1,183 enumeration seeds spanning 10 domains and 17 languages, with universes grounded in authoritative sources to ensure reproducibility. Representative open-source LLMs demonstrate severe limitations, achieving only 5.26--36.88 F1 on universe enumeration and 16.00--44.19 accuracy on knowledge-grounded reasoning. Diagnostic analyses reveal three stages of failure: \emph{completeness}, or missing knowledge; \emph{awareness}, or failure to identify requirements; and \emph{application}, or incorrect reasoning execution. This pattern persists across languages and model scales. Although test-time compute and retrieval augmentation yield measurable gains---up to 4.35 and 3.78 points, respectively---substantial gaps remain, exposing limitations in how current LLMs organize structured knowledge and execute compositional reasoning over bounded domains. The dataset is available at \url{https://huggingface.co/datasets/2npc/KnowledgeBerg}.
\end{abstract}

\section{Introduction}\label{sec:intro}

Large language models have become integral to applications spanning scientific research, industry, and everyday tasks \cite{raza2025industrial}. Despite the proliferation of benchmarks, most evaluations focus on isolated knowledge points, testing a single fact, definition, or concept \cite{yang-etal-2018-hotpotqa, hendryckstest2021, wang2024mmlu, zhang2025ontourlbenchmarkevaluatinglarge}, or on linear, step-by-step reasoning that evaluates a single chain of thought \cite{cobbe2021training, lightman2023let, lai-nissim-2024-mcot}. While these benchmarks are effective for measuring point-wise recall and basic reasoning, they rarely assess whether models can systematically cover entire knowledge domains and perform compositional reasoning across them.

This gap has significant implications in high-stakes domains such as education, healthcare, and life sciences, where success requires structured domain knowledge and reasoning rather than basic pattern matching. For instance, clinical diagnosis requires enumerating all plausible conditions before differential analysis; missing a critical candidate can lead to incorrect treatment decisions. More broadly, many seemingly simple questions conceal substantial latent requirements: a concise prompt may depend on a large, well-defined universe of entities, together with multi-step operations over that universe. We refer to this as the \emph{tip-of-the-iceberg} phenomenon.

\begin{figure}[t]
    \centering
    \includegraphics[width=\linewidth]{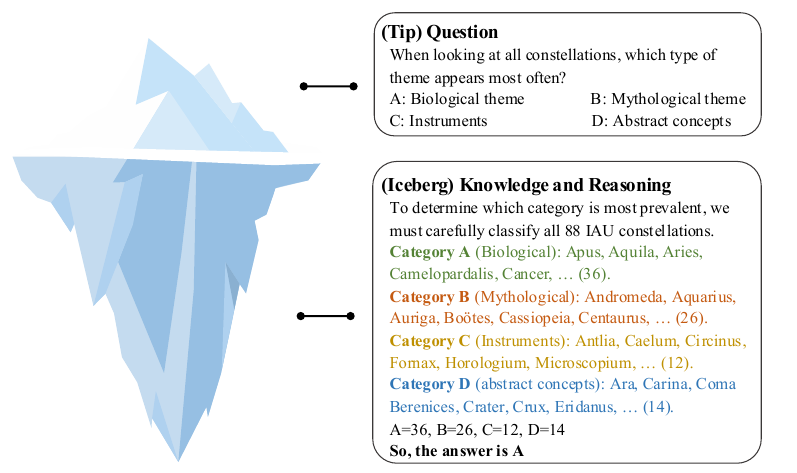}
    \caption{Illustration of the tip-of-the-iceberg phenomenon. A surface-simple question (tip) implicitly depends on a \emph{bounded universe} and \emph{compositional set-based reasoning} (submerged iceberg). In this example, answering requires knowledge of the 88 IAU-recognized constellations and a sequence of set operations: classify constellations, aggregate groups, count members, and compare cardinalities.}
    \label{fig:intro}
\end{figure}

Motivated by this limitation, we introduce a class of evaluation questions that combines knowledge coverage with compositional reasoning. Concretely, we characterize each question using two dimensions: \emph{knowledge width}, defined as the cardinality of the implicit universe required for a correct solution, and \emph{knowledge-grounded reasoning depth}, defined as the number of compositional set operations to derive the answer. This yields ``iceberg'' questions whose difficulty scales with width $\times$ depth. As illustrated in Figure~\ref{fig:intro}, answering a seemingly simple ``tip'' question about constellations requires access to the full constellation set and the ability to execute multiple operations, such as classification, aggregation, counting, and comparison. To make these latent universes explicit and reproducible, we curate bounded domain knowledge from verifiable sources, such as domain ontologies, textbooks, governmental documents and Wikipedia. Building on these universes, we design multiple families of questions that implicitly embed the required set, so that correct answers depend on both systematic knowledge coverage and set-based compositional reasoning. We call this benchmark \textsc{KnowledgeBerg}, inspired by the iceberg metaphor.

We compare \textsc{KnowledgeBerg} with existing benchmarks using \emph{Iceberg Gap}, which measures, across benchmarks, the gap between surface simplicity and latent width/depth demands. We further diagnose failure modes across representative LLMs and evaluate mitigation strategies on test-time computing. Overall, our contributions are:

\begin{itemize}[leftmargin=*,noitemsep,topsep=2pt]
    \item \textbf{Problem formulation and benchmark.} We formalize \emph{tip-of-the-iceberg} evaluation, where a concise question implicitly specifies a bounded universe and demands compositional set reasoning, and introduce \textsc{KnowledgeBerg}, a benchmark designed to jointly test bounded-universe coverage and set-based reasoning across 10 domains and 17 languages (4,800 questions) (\S\ref{sec:knowledgeberg}).
    \item \textbf{Cross-benchmark metric.} We propose \emph{Iceberg Gap}, a metric for comparing benchmarks by quantifying the gap between surface-form simplicity and latent requirements in knowledge width and reasoning depth (\S\ref{sec:metrics}).
    \item \textbf{Diagnostics and mitigation studies.} We provide large-scale evaluations and targeted analyses that localize failures into \emph{completeness--awareness--application}, and we quantify gains from representative approaches including test-time compute and retrieval augmentation (\S\ref{sec:diagnosis}, \S\ref{sec:improve}).
\end{itemize}

\section{Related Work}

\subsection{Knowledge Evaluation Benchmarks}

Most LLM benchmarks evaluate knowledge via isolated fact correctness. Exam-style benchmarks---MMLU \citep{hendryckstest2021}, MMLU-Pro \citep{wang2024mmlu}, MedQA \citep{jin2021disease}, and AI2 ARC \citep{clark2018think}---test factual recall across domains. Open-domain QA benchmarks such as TriviaQA \citep{joshi-etal-2017-triviaqa},
Natural Questions \citep{kwiatkowski-etal-2019-natural}, WebQuestions \citep{berant-etal-2013-semantic}, and PopQA \citep{mallen-etal-2023-trust}, together with knowledge-intensive suites such as KILT \citep{petroni-etal-2021-kilt}, evaluate answering with implicit or explicit evidence. Multi-step reasoning benchmarks---HotpotQA \citep{yang-etal-2018-hotpotqa}, GSM8K \citep{cobbe2021training}, and CommonsenseQA \citep{talmor-etal-2019-commonsenseqa}---stress inference chaining, but rarely require models to systematically cover a \emph{verifiably bounded} domain universe when forming an answer. \textsc{KnowledgeBerg} targets precisely this setting: questions whose answers require set-based composition over a complete, implicit bounded universe.

A smaller body of work evaluates \emph{coverage} or \emph{sufficiency}.
AmbigQA \citep{min-etal-2020-ambigqa} enumerates interpretations; ASQA
\citep{stelmakh-etal-2022-asqa} and ELI5 \citep{fan-etal-2019-eli5} assess aspect
coverage; and ProxyQA \citep{tan-etal-2024-proxyqa} verifies sufficiency. Related structured settings include knowledge-base and KG QA (ComplexWebQuestions \citep{talmor-berant-2018-web}, MetaQA \citep{zhang2017variational},
KQA Pro \citep{cao-etal-2022-kqa}) and compositional reasoning frameworks such as
Chameleon \citep{lu2023chameleon} and analyses of compositional failures in LLMs
\citep{li-etal-2024-understanding}.
These settings are closely related but typically expose the knowledge source, schema, or reasoning scaffold more explicitly. \textsc{KnowledgeBerg} differs by focusing on bounded-universe reasoning, where the question  gives no indication that complete-set coverage is required.

Recent work has also proposed benchmark-level notions of question difficulty.
Retrieval Complexity \citep{gabburo-etal-2024-measuring} characterizes how difficult it is
to retrieve the evidence needed to answer a question, whereas Iceberg Gap characterizes how
strongly a question conceals its universe-coverage and compositional requirements beneath a simple surface form. These perspectives are complementary: a question may be easy to
retrieve evidence for yet still exhibit a large Iceberg Gap if the prompt under-specifies
the need for complete-set coverage.

\subsection{Methods for Improving LLM Reasoning and Knowledge}

Prior work improves LLM reasoning reliability by eliciting and validating intermediate steps, including chain-of-thought prompting \citep{wei2022chain}, self-consistency \citep{wang2022selfconsistency}, multidimensional consistency \citep{lai2025multidimensionalconsistencyimprovesreasoning}, and self-verification and refinement \citep{weng-etal-2023-large,madaan2023self}. A complementary line augments parametric knowledge with external evidence via retrieval-augmented generation \citep{lewis2020retrieval}, with related interventions including query rewriting, decomposition, and abstraction before retrieval to partially externalize latent requirements and improve evidence acquisition \citep{chan2024rqraglearningrefinequeries,zheng2024take}.

\section{\textsc{KnowledgeBerg}}
\label{sec:knowledgeberg}
  
\subsection{Data Sources and Construction}
\label{sec:construction}
 
We manually curated structured knowledge spanning 10 domains (full taxonomy and
per-domain counts in Appendix~\ref{app:domains}) from five source types:
(i)~domain ontologies (e.g., Gene Ontology~\citep{ashburner2000gene},
NCBI Taxonomy~\citep{schoch2020ncbi});
(ii)~authoritative textbooks (e.g., \textit{Fundamentals of Physics}~\citep{halliday2013fundamentals});
(iii)~governmental and institutional documents (e.g., United Nations materials);
(iv)~official reference websites (e.g., the International Astronomical Union
and the National Basketball Association); and
(v)~Wikipedia as a supplementary source when canonical lists are stable and well-cited.
We exclude topics with factual disputes or lacking authoritative consensus
(e.g., territorial claims or rapidly evolving scientific hypotheses)
to ensure stability and reproducibility.
 
\paragraph{Bounded-set knowledge (EQ--EA).}
For each domain $d \in \mathcal{D}$, we construct enumeration questions in an
Enumeration-Question/Enumeration-Answer (EQ--EA) format.
Each enumeration question $\mathrm{EQ}_i$ is paired with a ground-truth answer set
$\mathrm{EA}_i = \{a_1, \ldots, a_{w_i}\}$,
where $w_i = |\mathrm{EA}_i|$ defines the \emph{knowledge width}.
We define the \emph{bounded universe} required by downstream reasoning as:
\begin{equation}
    U_i \;\triangleq\; \mathrm{EA}_i \;=\; \{a_1, \ldots, a_{w_i}\}.
\end{equation}
 
\paragraph{From knowledge to compositional reasoning.}
Building on EQ--EA pairs, we generate knowledge-grounded reasoning questions (KRQs)---multiple-choice questions with 8--10 options---that require models to reason over the bounded set $U_i$. Each KRQ $Q_j$ is instantiated by applying a short reasoning program $\mathbf{o}_j = \langle o_1, \ldots, o_{d_j} \rangle$ to a base enumeration question $\mathrm{EQ}_i$, where each step $o_t$ is drawn from a controlled set-operation schema including comparison, aggregation, constraint checking, filtering, counting, complementation, partitioning, ordering, union, and set equality. The program length $d_j = |\mathbf{o}_j|$ defines the reasoning depth of the KRQ.

\begin{equation}
    Q_j = \mathrm{apply}(\mathrm{EQ}_i,\; \mathbf{o}_j).
\end{equation}

We define \emph{reasoning depth} as the longest-path length of an operator graph.
For \textsc{KnowledgeBerg} items, the validated operator program is a linear chain
$\mathbf{o}_j = \langle o_1, \ldots, o_{d_j} \rangle$,
so the longest path of the operator graph reduces to the sequence length,
i.e., $\mathrm{Depth}(Q_j) = d_j = |\mathbf{o}_j|$.
For example:
\begin{equation}
    \text{Answer} = \mathrm{count}\!\left(\mathrm{filt}(U_i,\, \phi)\right),
\end{equation}
where $\phi$ is a predicate.
In practice, we use GPT-4o with structured prompts specifying $(EQ_i, o_j)$ to generate candidates, and validate them through a three-stage pipeline (review → adjudication → revision), in which two annotators independently assess candidate validity and disagreements are resolved through adjudication before final revision. Examples of EQ--KRQ pairs are provided in Appendix~\ref{app:examples_pairs}.
 
\paragraph{Iceberg structure.}
Each instance exposes a visible ``tip'' (the surface-level KRQ $Q_j$) while depending on a latent \emph{iceberg} characterized by knowledge width $w_i = |U_i|$ and reasoning depth $d_j = |\mathbf{o}_j|$.
 
\paragraph{Translations.}
We translate all 1{,}183 EQs and 4{,}800 KRQs \emph{from English} into 16 additional languages using Google Translate, and apply automatic sanity checks (language identification and script validation) to ensure format consistency.

\begin{table}[t]
\centering
\scriptsize
\setlength{\tabcolsep}{4pt}
\renewcommand{\arraystretch}{1.05}
\begin{tabularx}{\linewidth}{@{}X r@{}}
\toprule
\textbf{Category} & \textbf{Value} \\
\midrule
\multicolumn{2}{@{}l}{\textbf{Scale \& Coverage}} \\
Knowledge-grounded reasoning questions (KRQs) & 4,800 \\
Enumeration questions (EQs)                   & 1,183 \\
Domains                                        & 10    \\
Languages                                      & 17    \\
KRQs per EQ (avg)                             & 4.06  \\
Answer options (min / max)                    & 8 / 10 \\
\midrule
\multicolumn{2}{@{}l}{\textbf{Iceberg Structure}} \\
Knowledge width (avg / med / min / max) & 18.02 / 12 / 5 / 161 \\
Reasoning depth (avg / med / min / max) & 3.68 / 4 / 2 / 8     \\
\bottomrule
\end{tabularx}
\caption{Statistics of \textsc{KnowledgeBerg}.
Knowledge width is $|{\rm EA}_i|$.
Reasoning depth is the longest-path length of the operator graph;
for \textsc{KnowledgeBerg} items this equals $|\mathbf{o}_j|$,
since all validated programs are linear chains.}
\label{tab:data_stats}
\end{table}

\subsection{Dataset Statistics}
 
Table~\ref{tab:data_stats} summarizes \textsc{KnowledgeBerg}. The benchmark contains 4{,}800 KRQs derived from 1{,}183 EQs, covering 10 domains and 17 languages. On average, each EQ yields 4.06 KRQs through different operator sequences, ensuring diverse reasoning patterns. Each KRQ has 8--10 options with exactly one correct answer, corresponding to a random-guess baseline between 10.0\% and 12.5\%, depending on the item.

\subsection{Iceberg Gap: A Metric for Hidden Complexity}
\label{sec:metrics}

To quantify the ``tip versus mass'' contrast, we introduce \textbf{Iceberg Gap} (IG),
a score that is high when a question appears surface-simple yet implicitly requires
broad universe coverage and deep compositional reasoning.

\paragraph{Metric design.}
We represent each item as $(U, \Phi, q, a)$, where $U$ is the required bounded universe,
$\Phi$ the latent set-operation program, $q$ the surface question, and $a$ the correct answer.
In \textsc{KnowledgeBerg}, $U$ is the ground-truth set $\mathrm{EA}_i$; for benchmarks
without explicit universes, we estimate $|U|$ via a uniform two-step prompting procedure (domain identification $\to$ cardinality estimation).

IG comprises three components:
(1)~\textit{Surface Simplicity} $S$, computed from syntactic and semantic features;
(2)~\textit{Knowledge Width} $W$, with raw width $W_{\mathrm{raw}} = \log(1 + |U|)$ to reduce sensitivity to heavy-tailed universes; and (3)~\textit{Reasoning Depth} $D$, defined as the longest-path length of an operator graph
over a fixed operator vocabulary. We treat surface simplicity as an \emph{item property} rather than a model-specific surprisal measure, so that IG remains comparable across both open and closed models. For \textsc{KnowledgeBerg}, $\Phi$ is known by construction and $D$ reduces to the validated chain length; for external benchmarks, we estimate a minimal operator graph via a fixed LLM-based extraction procedure (Appendix~\ref{app:iceberg}). We acknowledge potential circularity in this external estimation and mitigate it by using the same prompt and model across all benchmarks, validating extracted depths on annotated subsets (Appendix~\ref{app:iceberg}).

To enable cross-benchmark comparison, we normalize $S$, $W$, and $D$ on a single pooled reference set $\Omega$ by mapping each raw value to its pooled percentile (mid-rank for ties), yielding scores in $(0, 1)$ (Appendix~\ref{app:iceberg}). Within-benchmark normalization is intentionally avoided: a raw width of $10$ may be extreme in a narrow benchmark but typical in a broader one, so benchmark-relative scaling
would undermine cross-dataset comparability. Percentile normalization preserves relative ordering while reducing sensitivity to heterogeneous scales and heavy-tailed distributions.

\paragraph{Aggregation.}
We combine the three components via a geometric mean:
\begin{equation}
    \mathrm{IG}(q) = \bigl(S(q)\cdot W(q)\cdot D(q)\bigr)^{1/3} \in (0,1).
\end{equation}
The geometric mean reflects the conjunctive nature of hidden complexity: a question
should score high only when it is \emph{simultaneously} surface-simple, broad in universe
coverage, and deep in reasoning. An additive aggregation would allow a single strong
component to mask deficits in the others, whereas the geometric mean penalizes imbalance.

\begin{figure}[t]
    \centering
    \includegraphics[width=\linewidth]{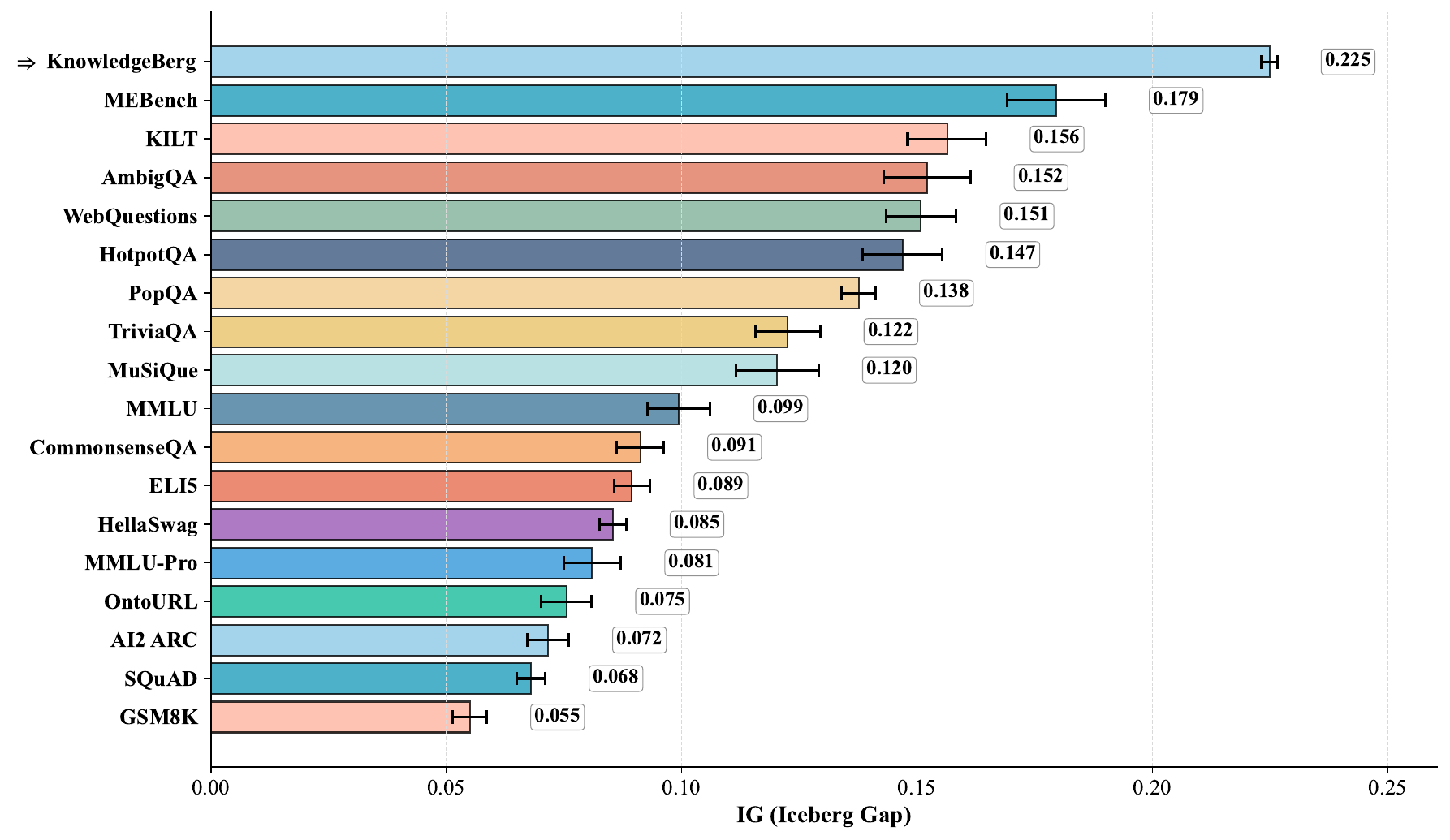}
    \caption{Iceberg Gap across benchmarks (mean $\pm$ 95\% CI),
    estimated from $N=500$ items sampled uniformly at random per benchmark.}
    \label{fig:gap}
\end{figure}

\paragraph{Benchmark comparison.}
Figure~\ref{fig:gap} reports IG as a descriptive characterization of hidden complexity
across benchmarks. \textsc{KnowledgeBerg} attains the highest IG ($0.225$), reflecting
the largest measured gap between surface simplicity and latent width/depth requirements.
Open-domain and multi-step QA benchmarks (WebQuestions, PopQA, KILT, HotpotQA) also
exhibit relatively high gaps, while exam-style and reading comprehension benchmarks
(MMLU, CommonsenseQA, SQuAD) score lower, indicating weaker joint pressure on
bounded-set coverage and set-based composition. GSM8K attains the lowest IG ($0.055$).

\section{Diagnosing LLMs on KnowledgeBerg}
\label{sec:diagnosis}

In this section, we evaluate representative LLMs on \textsc{KnowledgeBerg} to diagnose where failures arise in the knowledge-to-answer process. We first report overall performance and show that enumeration completeness and KRQ accuracy are only weakly coupled. Motivated by this gap, we introduce a three-stage diagnostic framework: \textbf{completeness}, \textbf{awareness}, and \textbf{application}. We examine this framework from two complementary angles: correlational analyses of structural difficulty factors (\S\ref{subsec:difficulty_factors}) and controlled inference-time prompt variants that selectively provide knowledge or encourage reasoning to more directly stress each stage (\S\ref{subsec:diagnosis}). We then validate the findings cross-lingually (\S\ref{subsec:multilingual}).

\subsection{Settings}
\label{sec:kb_settings}

We evaluate representative open-source LLMs from the Qwen~\citep{yang2025qwen3}, LLaMA~\citep{dubey2024llama}, Mistral, Phi, and Gemma~\citep{team2025gemma} families on \textsc{KnowledgeBerg}. Models answer enumeration questions (EQs) and knowledge-grounded reasoning questions (KRQs) using a unified zero-shot prompt with greedy decoding (temperature $=0.0$, top-$p=1.0$; Appendix~\ref{app:exp_setup}).

We report Universe F1 for EQs and accuracy for KRQs. For KRQs, we deterministically extract the final answer option from the \texttt{\textbackslash boxed\{\}} span using a regular expression and count unparseable outputs as incorrect. For EQs, we compute set-level Universe F1 using a hybrid protocol: we first apply rule-based normalization and matching, and invoke an LLM judge (Qwen3-30B-A3B-Instruct-2507; \citealp{yang2025qwen3}) only for the remaining unmatched cases (Appendix~\ref{app:llm_judge_retrieval}).

\subsection{Overall Performance: A Puzzling Ceiling}
\label{subsec:main_results}

\begin{table}[t]
\centering
\small
\setlength{\tabcolsep}{4pt}
\renewcommand{\arraystretch}{0.95}
\begin{tabular}{lcc}
\toprule
\textbf{Model} & \textbf{Universe F1} & \textbf{KRQ Acc.} \\
\midrule
Qwen3.5-0.8B                    &  5.26 & 24.38 \\
Qwen3.5-9B                      & 11.60 & 36.35 \\
Qwen3.5-27B                     & 12.16 & 44.19 \\
Qwen3.5-35B-A3B                 & 14.29 & 41.71 \\
Mistral-Small-24B-Instruct-2501 & 33.26 & 19.88 \\
Phi-4-mini-instruct             &  9.75 & 16.00 \\
Phi-4-14B                           & 25.17 & 38.79 \\
Llama-3.3-70B-Instruct          & 36.88 & 35.90 \\
Gemma-3-4B-it                   & 16.51 & 29.42 \\
Gemma-3-12B-it                  & 23.82 & 31.06 \\
Gemma-3-27B-it                  & 27.32 & 32.81 \\
\bottomrule
\end{tabular}
\caption{Performance on \textsc{KnowledgeBerg} (English). Universe F1 measures completeness on EQs, and KRQ accuracy measures knowledge-grounded reasoning.}
\label{tab:main_results}
\end{table}

Table~\ref{tab:main_results} reveals two patterns across models of varying architectures and scales. First, completeness on EQs remains consistently limited. Universe F1 ranges from 5.26 to 36.88, with most models remaining below 30 and only Llama-3.3-70B-Instruct and Mistral-Small-24B-Instruct-2501 exceeding that level. This indicates that even when the relevant knowledge is constrained to a bounded, structured set, faithfully enumerating it remains difficult. Second, KRQ accuracy also remains modest overall, ranging from 16.00 to 44.19 despite the multiple-choice format of KRQs, which should in principle reduce the search space.

More importantly, the two metrics are not aligned. Some models with relatively strong enumeration completeness perform poorly on KRQs; for example, Mistral-Small-24B reaches 33.26 Universe F1 but only 19.88 KRQ accuracy. Conversely, some models with limited completeness still achieve comparatively strong KRQ performance: Qwen3.5-27B attains only 12.16 Universe F1 yet reaches the highest KRQ accuracy at 44.19. This mismatch suggests that bounded-universe coverage alone is insufficient to explain model behavior on \textsc{KnowledgeBerg}.

\begin{figure}[t]
\centering
\includegraphics[width=\linewidth]{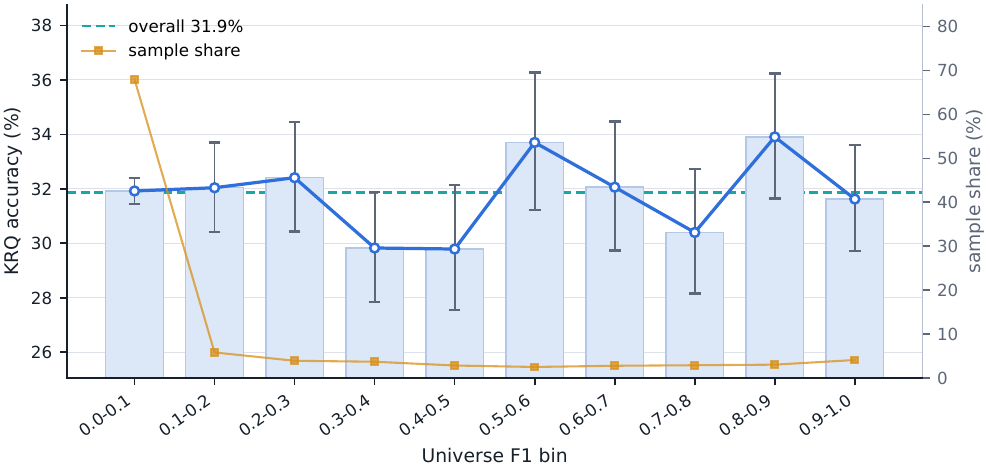}
\caption{KRQ accuracy versus enumeration quality (Universe F1), with instances binned by F1 and aggregated across models. Instance-level correlations are near-zero (Spearman $\rho=0.0023$; Kendall $\tau=0.0020$).}
\label{fig:retrieval_mcq}
\end{figure}

\paragraph{The Decoupling Puzzle.}

The weak coupling between EQ and KRQ performance becomes even clearer at the instance level. Intuitively, higher Universe F1 should translate into higher KRQ accuracy: if more of the required knowledge is available, the model should reason more reliably. However, Figure~\ref{fig:retrieval_mcq} shows that this relationship is essentially absent. After binning instances by Universe F1 and aggregating across models, KRQ accuracy varies only weakly and non-monotonically across bins, and the instance-level correlations are near zero (Spearman $\rho=0.0023$; Kendall $\tau=0.0020$). Thus, limited knowledge coverage is not the only source of failure: even when a model recovers more of the relevant universe, it may still fail to convert that knowledge into the correct answer.

\paragraph{A Three-Stage Hypothesis.}

This decoupling suggests that failures extend beyond knowledge coverage alone. Even when relevant facts are available, a model may still fail in two additional ways: it may not identify the question's implicit knowledge requirements, or it may execute the required reasoning operations incorrectly. We therefore decompose the knowledge-to-answer process into three stages: \textbf{completeness} (is the required universe knowledge available in the model?), \textbf{awareness} (does the model correctly identify what knowledge the question requires?), and \textbf{application} (does it correctly apply the identified knowledge through the necessary reasoning operations to arrive at the correct answer?). We examine this hypothesis in two ways: analyses of how accuracy varies with structural difficulty factors (\S\ref{subsec:difficulty_factors}) and controlled inference-time prompt variants that test whether providing knowledge or structuring reasoning can isolate each stage's contribution (\S\ref{subsec:diagnosis}).

\subsection{Correlational Analysis: Structural Difficulty Factors}
\label{subsec:difficulty_factors}

\begin{figure}[t]
\centering
\includegraphics[width=\linewidth]{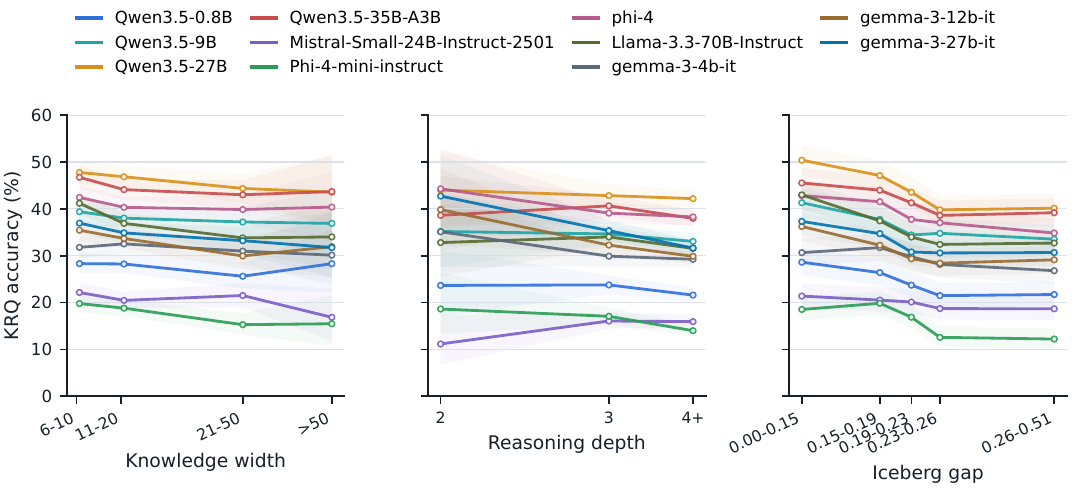}
\caption{KRQ accuracy across bins of knowledge width, reasoning depth, and Iceberg Gap (English). Each line denotes one model.}
\label{fig:difficulty_factors}
\end{figure}

To localize where failures arise, we analyze how KRQ accuracy varies with three structural properties of questions (Figure~\ref{fig:difficulty_factors}). \textbf{Knowledge width} (the number of required universe elements) captures the size of the knowledge space and is most closely associated with \emph{completeness}. \textbf{Reasoning depth} (the number of reasoning steps) captures the length of the inference chain over that universe and is most closely associated with \emph{application}. \textbf{Iceberg Gap (IG)} captures the mismatch between surface simplicity and hidden complexity, and is intended to probe \emph{awareness}: whether models recognize the extensive latent requirements beneath concise questions.

\paragraph{Structural patterns reveal stage-associated bottlenecks.}

Figure~\ref{fig:difficulty_factors} shows a consistent qualitative pattern across all three axes. As knowledge width increases from 6--10 to $>50$ elements, KRQ accuracy generally declines across models, with the largest-width bucket being the most challenging. This pattern is consistent with \emph{completeness}-related bottlenecks: larger universes are harder to recover exhaustively, which aligns with the low Universe F1 values in Table~\ref{tab:main_results}. Accuracy also tends to decrease as reasoning depth increases from 2 to 4+, suggesting more frequent \emph{application}-related failures in longer compositional chains. Finally, higher IG bins are associated with systematically lower KRQ accuracy, which is compatible with persistent \emph{awareness}-related failures: when a question appears surface-simple but conceals broader latent requirements, models often fail to recognize what the task actually demands.

\subsection{Diagnostic Prompt Variants}
\label{subsec:diagnosis}

\begin{figure}[t]
    \centering
    \includegraphics[width=\linewidth]{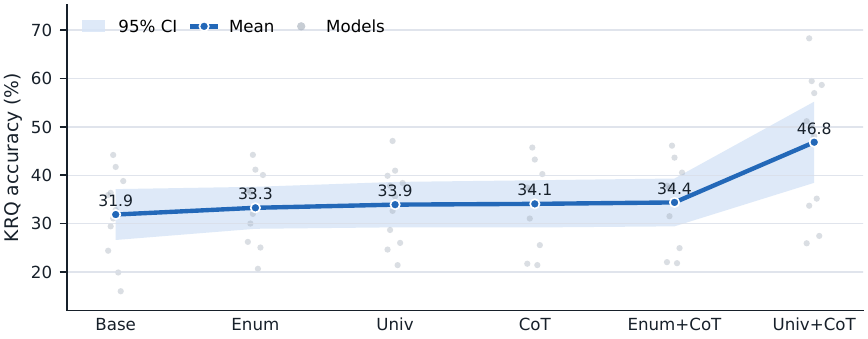}
    \caption{Effect of diagnostic prompt variants on KRQ accuracy (English). Thin lines show individual models; the highlighted curve shows the mean across models, with 95\% confidence intervals.}
    \label{fig:diagnosis}
\end{figure}

While informative, the correlational analyses in \S\ref{subsec:difficulty_factors} do not by themselves isolate causal mechanisms. We therefore conduct controlled prompt interventions that selectively manipulate \emph{universe knowledge availability} and \emph{reasoning scaffolding} to probe which stages of the knowledge-to-answer process are most sensitive to these changes (Figure~\ref{fig:diagnosis}).

\paragraph{Prompt variants.}

We compare \textsc{Base} (the default KRQ prompt) with five variants. \textsc{Univ} provides the gold universe, allowing us to probe how strongly performance depends on access to the complete bounded set. \textsc{Enum} asks the model to list the \emph{candidate} universe elements it considers relevant \emph{before} choosing an option, aiming to externalize evidence selection and probe whether making candidate knowledge explicit helps with requirement identification. \textsc{CoT} encourages step-by-step reasoning to probe whether structured reasoning reduces errors in executing the required operations. Finally, \textsc{Enum+CoT} and \textsc{Univ+CoT} examine whether these interventions interact.

\paragraph{Single interventions help only modestly; combining knowledge and reasoning helps substantially more.}

Figure~\ref{fig:diagnosis} shows that all single interventions yield only limited gains over the \textsc{Base} prompt on the model average. Mean KRQ accuracy rises from 31.9 under \textsc{Base} to 33.3 with \textsc{Enum}, 33.9 with \textsc{Univ}, and 34.1 with \textsc{CoT}; \textsc{Enum+CoT} reaches 34.4. By contrast, \textsc{Univ+CoT} yields a much larger improvement, reaching 46.8.

Two observations follow. First, neither externalizing candidate knowledge nor applying a single intervention in isolation is sufficient to address the benchmark. Second, the strongest gains arise when complete universe knowledge is paired with explicit reasoning scaffolding, suggesting that missing knowledge and errors in reasoning execution interact in practice rather than behaving as fully separable sources of failure.

\paragraph{A complementary error analysis of strong closed-source models.}

Although \textsc{Univ+CoT} yields the strongest improvement among our prompt variants, performance remains well below ceiling for open-source models. To obtain a complementary view of residual failures under a stronger model regime, we additionally evaluate two closed-source models under \textsc{CoT}: \textsc{DeepSeek-Chat} achieves 48.39\%, and \textsc{Gemini-3-Flash} reaches 65.24\%. We then manually inspect 100 sampled \textsc{Gemini-3-Flash} errors (Appendix~\ref{app:error_study}), finding \textbf{completeness}-related errors in 42\% of cases, \textbf{application}-related errors in 38\%, \textbf{awareness}-related errors in 17\%, and \textbf{artifacts} in 3\%.

Because this manual study is conducted under \textsc{CoT} rather than \textsc{Univ+CoT}, we treat it as complementary evidence rather than as a direct decomposition of the residual gap under the strongest prompt condition. Even so, the observed distribution is broadly consistent with the prompt-variant results: completeness- and application-related failures remain dominant, whereas the limited gains from \textsc{Enum} suggest that simply eliciting candidate lists does not reliably resolve requirement-identification failures.

\begin{table}[t]
\centering
\scriptsize
\setlength{\tabcolsep}{3pt}
\renewcommand{\arraystretch}{1.0}
\begin{tabular*}{\linewidth}{@{\extracolsep{\fill}}lrrrrr}
\toprule
\textbf{Model} & \textbf{All} & \textbf{High} & \textbf{Mid} & \textbf{Low} & \textbf{Std.} \\
\midrule
Qwen3.5-0.8B                         & 23.01 & 23.88 & 22.17 & 19.05 &  3.08 \\
Qwen3.5-9B                           & 31.22 & 32.07 & 29.66 & 28.42 &  2.40 \\
Qwen3.5-27B                          & \textbf{38.33} & \textbf{39.07} & \textbf{36.78} & \textbf{36.19} &  2.43 \\
Qwen3.5-35B-A3B                      & 36.62 & 37.17 & 35.52 & 34.97 &  1.65 \\
Mistral-Small-24B-Instruct-2501      & 23.32 & 23.17 & 24.17 & 23.00 &  5.32 \\
Phi-4-mini-instruct                  & 15.91 & 16.00 & 15.11 & 16.58 & \textbf{1.04} \\
Phi-4-14B                                & 32.85 & 34.53 & 30.38 & 26.46 &  3.22 \\
Llama-3.3-70B-Instruct               & 30.63 & 31.87 & 26.73 & 29.03 &  3.36 \\
Gemma-3-4B-it                        & 27.08 & 27.26 & 26.84 & 26.32 &  1.54 \\
Gemma-3-12B-it                       & 28.73 & 28.94 & 29.15 & 26.84 &  1.20 \\
Gemma-3-27B-it                       & 30.65 & 31.02 & 30.26 & 29.02 &  1.18 \\
\bottomrule
\end{tabular*}
\caption{Multilingual KRQ accuracy (\%) by computational resource tier. High includes Arabic, Chinese, Dutch, English, French, German, Hindi, Italian, Japanese, Korean, Russian, and Spanish; Mid includes Bengali, Thai, and Turkish; Low includes Swahili and Telugu. Std. indicates cross-lingual consistency.}
\label{tab:multilingual}
\end{table}

\subsection{Cross-Lingual Validation}
\label{subsec:multilingual}

To test whether our findings generalize beyond English, we evaluate models on \textsc{KnowledgeBerg}'s multilingual partition.

Table~\ref{tab:multilingual} shows that the overall ranking is broadly preserved across languages: Qwen3.5-27B remains the strongest model overall (38.33) and in each resource tier. Performance typically degrades from high- to low-resource languages, but the magnitude of the drop is usually modest rather than catastrophic. The largest degradation is observed for Phi-4-14B (34.53 $\rightarrow$ 26.46), while several models remain comparatively stable across tiers; for example, Gemma-3-27B varies only from 31.02 to 29.02, and Phi-4-mini-instruct has the lowest cross-lingual variance overall (Std.\ 1.04). By contrast, Mistral-Small-24B-Instruct-2501 is the least stable model in this set (Std.\ 5.32).

\begin{table}[t]
\centering
\scriptsize
\setlength{\tabcolsep}{2.2pt}
\renewcommand{\arraystretch}{0.95}
\begin{tabular}{lccccc}
\toprule
 & $\Delta$E & $\Delta$C & $\Delta$U & $\Delta$E{+}C & $\Delta_{\mathrm{C}\mid \mathrm{U}}$ \\
\midrule
Avg.  & +1.8 & +3.1 & +2.8 & +3.5 & +13.6 \\
Range & [+0.6, +3.0] & [+1.8, +4.6] & [+1.5, +4.2] & [+2.0, +5.0] & [+10.2, +16.9] \\
\bottomrule
\end{tabular}
\caption{Cross-lingual gains of diagnostic prompt variants, measured in accuracy points. Range is computed across languages. $\Delta$E, $\Delta$C, and $\Delta$U denote gains of \textsc{Enum}, \textsc{CoT}, and \textsc{Univ} over \textsc{Base}; $\Delta$E{+}C denotes the gain of \textsc{Enum+CoT} over \textsc{Base}; and $\Delta_{\mathrm{C}\mid \mathrm{U}}$ denotes $(\textsc{Univ+CoT})-\textsc{Univ}$.}
\label{tab:crosslingual_prompt_effects}
\end{table}

\begin{table*}[t]
\centering
\footnotesize
\setlength{\tabcolsep}{3.5pt}
\renewcommand{\arraystretch}{0.98}
\begin{tabular}{l c ccc ccc ccc ccc}
\toprule
\multirow{2}{*}{\textbf{Model}} &
\multirow{2}{*}{\textbf{Base}} &
\multicolumn{3}{c}{\textbf{Self-Consistency}} &
\multicolumn{3}{c}{\textbf{Self-Refine}} &
\multicolumn{3}{c}{\textbf{Self-Verifier}} &
\multicolumn{3}{c}{\textbf{RAG}} \\
\cmidrule(lr){3-5}\cmidrule(lr){6-8}\cmidrule(lr){9-11}\cmidrule(lr){12-14}
& &
\textbf{sc=4} & \textbf{sc=8} & \textbf{sc=16} &
\textbf{k=1} & \textbf{k=2} & \textbf{k=3} &
\textbf{n=4} & \textbf{n=8} & \textbf{n=16} &
\textbf{dense} & \textbf{hybrid} & \textbf{rerank} \\
\midrule
Qwen3.5-0.8B                    & 24.38 & 25.76 & 26.54 & \best{27.18} & 24.72 & 21.64 & 20.98 & 25.88 & 26.42 & \second{26.88} & 26.34 & 26.78 & 26.61 \\
Qwen3.5-9B                      & 36.35 & 37.82 & 38.34 & \best{39.05} & 36.71 & 33.02 & 32.41 & 37.92 & 38.41 & \second{38.95} & 38.37 & 38.74 & 38.58 \\
Qwen3.5-27B                     & 44.19 & 45.72 & 46.31 & \best{47.19} & 44.08 & 40.76 & 39.91 & 45.84 & 46.38 & \second{46.89} & 46.42 & 46.69 & 46.55 \\
Qwen3.5-35B-A3B                 & 41.71 & 43.14 & 43.88 & \best{44.78} & 41.52 & 37.95 & 37.22 & 43.21 & 43.91 & \second{44.46} & 43.88 & 44.18 & 44.02 \\
Mistral-Small-24B & 19.88 & 21.91 & 23.05 & \best{23.98} & 19.74 & 16.42 & 15.83 & 22.03 & 23.16 & \second{23.78} & 22.84 & 23.28 & 23.05 \\
Phi-4-mini-instruct             & 16.00 & 18.22 & 19.08 & \best{20.35} & 16.38 & 13.85 & 13.12 & 18.35 & 19.42 & \second{20.06} & 19.31 & 19.78 & 19.56 \\
Phi-4-14B                           & 38.79 & 40.18 & 40.92 & \best{41.68} & 38.62 & 35.41 & 34.78 & 40.24 & 40.88 & \second{41.42} & 40.84 & 41.17 & 41.03 \\
Llama-3.3-70B-Instruct          & 35.90 & 37.42 & 38.13 & \best{38.91} & 36.04 & 33.18 & 32.76 & 37.55 & 38.20 & \second{38.71} & 38.09 & 38.45 & 38.28 \\
Gemma-3-4B-it                   & 29.42 & 31.08 & 31.74 & \best{33.12} & 29.66 & 26.31 & 25.70 & 31.22 & 31.88 & \second{32.43} & 31.73 & 32.10 & 31.91 \\
Gemma-3-12B-it                  & 31.06 & 33.02 & 33.74 & \best{35.11} & 31.58 & 28.27 & 27.35 & 33.16 & 34.02 & \second{34.72} & 34.31 & 34.62 & 34.48 \\
Gemma-3-27B-it                  & 32.81 & 34.42 & 35.08 & \best{36.29} & 33.04 & 30.12 & 29.58 & 34.55 & 35.24 & \second{35.86} & 35.44 & 35.72 & 35.61 \\
\bottomrule
\end{tabular}
\caption{Testing-time enhancements for KRQ across models. Boldface indicates the best configuration per model; underlined indicates the second best.}
\label{tab:enhance_across_models}
\end{table*}

Applying the same prompt variants across languages (Table~\ref{tab:crosslingual_prompt_effects}) reveals the same qualitative pattern as in English. \textsc{Enum} again has the smallest effect, yielding an average gain of +1.8 points. \textsc{CoT} and \textsc{Univ} each provide somewhat larger but still modest improvements (+3.1 and +2.8 points on average, respectively), while \textsc{Enum+CoT} reaches +3.5 points. By far the largest gain arises when \textsc{CoT} is added on top of \textsc{Univ}: $\Delta_{\mathrm{C}\mid \mathrm{U}}$ averages +13.6 points across languages, with gains ranging from +10.2 to +16.9. Taken together, these multilingual results support the same overall diagnosis as in English: explicit candidate enumeration contributes comparatively little on its own, whereas the strongest improvements emerge when complete universe knowledge is paired with reasoning scaffolding.

\section{Improve LLMs on KnowledgeBerg}
\label{sec:improve}

We evaluate four families of testing-time strategies. \textbf{Self-consistency} samples multiple trajectories and aggregates them by majority vote, primarily targeting instability in \emph{application}. \textbf{Self-refinement} alternates \textit{Critique} and \textit{Revise} to iteratively edit solutions using self-generated feedback, but may become counterproductive when critiques reinforce incorrect premises. \textbf{Self-verification} decouples proposal from selection: the model first generates candidate solutions and then uses a verifier to select the most plausible answer, aiming to improve final option selection. \textbf{Retrieval-augmented generation (RAG)} adds retrieved passages to the input to alleviate \emph{completeness} gaps. We compare dense retrieval, hybrid retrieval (BM25+dense), and cross-encoder reranking. Our retrieval corpus is a lightweight collection derived from dataset construction (ontologies, curated sources, and official documents); we intentionally avoid large general-purpose databases (e.g., Wikidata) to keep retrieval effects interpretable and enable ablations.

Table~\ref{tab:enhance_across_models} reveals a clear ranking among testing-time strategies. \textbf{Self-consistency} is the strongest and most reliable intervention: performance increases monotonically with more samples, and \texttt{sc=16} is the best configuration for all 11 models. This pattern suggests that a substantial portion of KRQ errors stems from inference-time instability in executing the required reasoning operations. \textbf{Self-verification} is consistently the next strongest method: performance also improves monotonically with verifier budget, and \texttt{n=16} is the second-best configuration for all 11 models.

\textbf{RAG} provides robust but smaller improvements. All three retrieval variants outperform the base model for all 11 models, and hybrid retrieval is consistently the strongest RAG configuration. This result suggests that retrieval quality matters more than simply appending external passages: combining sparse and dense retrieval yields more useful evidence than dense retrieval alone, while reranking offers only limited additional benefit. Finally, \textbf{self-refinement} is the least effective strategy. A single refinement step yields only marginal gains on a subset of models, whereas additional refinement rounds consistently reduce accuracy, often substantially. This pattern is consistent with the diagnosis in \S\ref{sec:diagnosis}: when the initial reasoning trajectory is grounded in incomplete or mistaken universe knowledge, iterative self-critique may amplify those errors rather than correct them.

\section{Conclusion}
\textsc{KnowledgeBerg} highlights a limitation of current large language models that is easy to miss in conventional evaluations: models may answer surface-simple questions poorly not because the questions are linguistically difficult, but because they require both systematic coverage of a bounded knowledge universe and reliable reasoning over that universe. Across our experiments, current models remain far from robust in this setting, and the gap cannot be explained by missing knowledge alone. Instead, our analyses point to failures at multiple stages of the knowledge-to-answer process, including completeness, awareness, and application. Inference-time interventions improve performance, but they do not remove the underlying difficulty. We therefore view \textsc{KnowledgeBerg} not only as a benchmark, but also as a testbed for studying how language models organize structured knowledge and reason over explicit sets---capabilities that are especially important in high-stakes domains.

\section{Limitations}

\paragraph{Universe Construction and Domain Coverage.}
Although universes are grounded in authoritative sources, the choice of 10 domains and 1,183 seeds is manually curated and may introduce sampling bias. We also exclude fast-changing topics for stability, so the benchmark may underrepresent dynamic knowledge domains. Finally, most universes are small (median size 12), and very large universes (e.g., thousands of entities) are limited.

\paragraph{Evaluation Protocol Dependencies.}
Universe F1 uses rule-based normalization plus an LLM judge for unresolved matches, which can introduce model-dependent bias. Our operator set for reasoning depth is also a simplification, and depth extraction for external benchmarks relies on LLM inference, posing a risk of circular reasoning despite standardized prompting and validation.

\paragraph{Multilingual Evaluation Scope.}
The 17-language setting relies on machine translation with automatic checks but lacks comprehensive native-speaker validation. Translation artifacts may change difficulty or introduce ambiguity, especially for low-resource languages, and we do not explicitly analyze cultural/linguistic variation in how knowledge is organized or enumerated.

\bibliography{custom}

\clearpage

\appendix

\section{Details of \textsc{KnowledgeBerg}}\label{app:kb_details}

\subsection{Domains}\label{app:domains}

Table~\ref{tab:domains} lists the 10 domains in \textsc{KnowledgeBerg}.

\begin{table}[h]
\centering
\scriptsize
\setlength{\tabcolsep}{5pt}
\renewcommand{\arraystretch}{1.05}
\begin{tabularx}{\linewidth}{r X r r}
\toprule
\textbf{ID} & \textbf{Domain} & \textbf{\#EQs} & \textbf{\#KRQs} \\
\midrule
01 & Science, Technology \& Computing & 130 & 525 \\
02 & Life Science, Health \& Food & 311 & 1,234 \\
03 & Humanities, Culture \& Language & 200 & 794 \\
04 & Geography \& Geopolitics & 105 & 434 \\
05 & Economy, Finance \& Industry & 98 & 457 \\
06 & Time \& Calendar & 59 & 228 \\
07 & Sports & 75 & 295 \\
08 & Environment \& Climate & 67 & 259 \\
09 & Transport \& Infrastructure & 58 & 248 \\
10 & Legal \& Public Policy & 80 & 326 \\
\midrule
\textbf{All} & \textbf{Total} & \textbf{1,183} & \textbf{4,800} \\
\bottomrule
\end{tabularx}
\caption{Domain taxonomy and per-domain counts of enumeration questions (EQs) and knowledge-grounded reasoning questions (KRQs).}
\label{tab:domains}
\end{table}

\subsection{Example EQ--EA and Derived KRQs}\label{app:examples_pairs}

As shown in Table~\ref{tab:ex_pair_life_science} and \ref{tab:ex_pair_transport}, we provide representative EQ--EA seeds and their derived KRQs to illustrate how bounded universes and operator sequences are instantiated in concrete multiple-choice questions.

\begin{table}[h]
\centering
\scriptsize
\setlength{\tabcolsep}{3pt}
\renewcommand{\arraystretch}{0.98}
\begin{tabularx}{\columnwidth}{@{}X@{}}
\toprule
\textbf{Pair 1 (Life Science, Health \& Food).} \\
\midrule
\textbf{Enumeration seed (EQ).}
Which parts of the alimentary tract are listed by the National Cancer Institute SEER Training materials? \\
\textbf{Enumeration answer (EA).}
Mouth; Pharynx; Esophagus; Stomach; Small intestine; Large intestine; Rectum; Anus. \\
\addlinespace[2pt]
\textbf{Derived iceberg KRQ (Operation: Ordering + Comparison).}
Using that validated SEER alimentary-tract closed set, compare the items before the stomach with the items after the stomach. \\
\textbf{Options.}
\begin{enumerate}[label=\Alph*., leftmargin=1.4em, itemsep=0pt, topsep=2pt, parsep=0pt]
\item The pre-stomach subset is larger by exactly 1.
\item The post-stomach subset is larger by exactly 1.
\item The two subsets tie.
\item The pre-stomach subset is larger by exactly 2.
\item The post-stomach subset is larger by exactly 2.
\item The pre-stomach subset has only 2 items.
\item The post-stomach subset has 5 items.
\item Together the two subsets account for only 6 items.
\end{enumerate}
\textbf{Gold.} \textbf{B} \\
\bottomrule
\end{tabularx}
\caption{Seed-aligned EQ--EA and a derived KRQ.}
\label{tab:ex_pair_life_science}
\end{table}

\begin{table}[h]
\centering
\scriptsize
\setlength{\tabcolsep}{3pt}
\renewcommand{\arraystretch}{0.98}
\begin{tabularx}{\columnwidth}{@{}X@{}}
\toprule
\textbf{Pair 2 (Transport \& Infrastructure).} \\
\midrule
\textbf{Enumeration seed (EQ).}
Which important parts and terms are involved in Indian Railways switch assembly terminology? \\
\textbf{Enumeration answer (EA).}
Stock rail joint; Stock rail; Tongue rail; Heel of switch; Switch angle at toe; Divergence at heel; Heel block; Throw at toe; Length of switch; Bolts. \\
\addlinespace[2pt]
\textbf{Derived iceberg KRQ (Operation: Complement + Comparison).}
Within the validated Indian Railways switch-assembly terminology set, compare the pure-rail subset with its complement.
The pure-rail subset is stock rail and tongue rail. Which conclusion is correct? \\
\textbf{Options.}
\begin{enumerate}[label=\Alph*., leftmargin=1.4em, itemsep=0pt, topsep=2pt, parsep=0pt]
\item The pure-rail subset is larger by exactly 5.
\item The pure-rail subset is larger by exactly 6.
\item The two sides are equal in size.
\item The complement is larger by exactly 4.
\item The complement is larger by exactly 5.
\item The complement is larger by exactly 6.
\item The pure-rail subset is smaller by exactly 8.
\item The complement is larger by exactly 7.
\end{enumerate}
\textbf{Gold.} \textbf{F} \\
\bottomrule
\end{tabularx}
\caption{Seed-aligned EQ--EA and a derived KRQ.}
\label{tab:ex_pair_transport}
\end{table}

\subsection{Languages}\label{app:languages}
We translate all questions from English into 16 additional languages using Google Translate. We categorize the 17 languages into three computational resource tiers based on their availability in contemporary LLM pretraining corpora: High-resource (Arabic, Chinese, Dutch, English, French, German, Hindi, Italian, Japanese, Korean, Russian, and Spanish), Mid-resource (Bengali, Thai, and Turkish), and Low-resource (Swahili and Telugu).

\section{Iceberg Gap: Implementation Details}
\label{app:iceberg}

This appendix summarizes the details needed to reproduce Iceberg Gap (IG). We describe (i) how we compute surface simplicity and its corresponding surface complexity, (ii) how we define raw width and depth, and (iii) how we normalize all three components using a pooled percentile transformation.

\subsection{Surface Simplicity and Surface Complexity}
\label{app:surface_complexity}

We first compute a bounded \textbf{surface simplicity} score $S_{\mathrm{surf}}(q)\in[0,1]$ from syntactic and semantic factors, and then convert it into a bounded \textbf{surface complexity} value $SC(q)\in[0,SC_{\max}]$.

\paragraph{Syntactic simplicity.}
Let $n$ be the number of tokens in $q$, $\bar{s}$ the average sentence length (in tokens), and $p$ the punctuation density (punctuation count divided by total characters). We define
\begin{align}
S_{\text{len}}(q)&=1-\min\!\left(1,\frac{n}{L_{\max}}\right),\\
S_{\text{sent}}(q)&=1-\min\!\left(1,\frac{\bar{s}}{S_{\max}}\right),\\
S_{\text{punct}}(q)&=1-\min\!\left(1,\frac{p}{P_{\max}}\right),
\end{align}
with fixed caps $L_{\max}=40$, $S_{\max}=30$, and $P_{\max}=0.08$. We then average the three terms:
\begin{equation}
S_{\text{syn}}(q)=\tfrac{1}{3}\big(S_{\text{len}}(q)+S_{\text{sent}}(q)+S_{\text{punct}}(q)\big).
\end{equation}

\paragraph{Semantic simplicity.}
We compute GPT-2 perplexity $\mathrm{PP}(q)$ and normalize it with a fixed cap $\mathrm{PP}_{\max}=1000$:
\begin{equation}
S_{\text{sem}}(q)=1-\min\!\left(1,\frac{\mathrm{PP}(q)}{\mathrm{PP}_{\max}}\right).
\end{equation}

\paragraph{Combining the two.}
We combine syntactic and semantic simplicity with $\alpha=0.5$:
\begin{equation}
S_{\mathrm{surf}}(q)=(1-\alpha)S_{\text{syn}}(q)+\alpha S_{\text{sem}}(q).
\end{equation}
We then convert surface simplicity into surface complexity:
\begin{equation}
SC(q)=SC_{\max}\bigl(1-S_{\mathrm{surf}}(q)\bigr),
\end{equation}
with $SC_{\max}=30$. Equivalently, the raw simplicity component used by IG is
\begin{equation}
S_{\mathrm{raw}}(q)=1-\frac{\mathrm{clip}(SC(q),0,SC_{\max})}{SC_{\max}}=S_{\mathrm{surf}}(q),
\end{equation}
up to clipping at the fixed cap.

\subsection{Raw Width/Depth and Pooled Percentile}
\label{app:pooled_ecdf}

\paragraph{Raw width and depth.}
We define raw knowledge width as
\begin{equation}
W_{\mathrm{raw}}(q)=\log(1+|U|),
\label{eq:w_raw}
\end{equation}
which reduces sensitivity to heavy-tailed universe sizes. We let $D_{\mathrm{raw}}(q)$ denote reasoning depth: the annotated depth when available, and otherwise the longest-path length of an operator DAG extracted from the question. When $|U|$ or the operator DAG is not explicitly available, we estimate them with LLM prompts (Tables~\ref{tab:llm_prompt_width} and~\ref{tab:llm_prompt_dag}).

\begin{table}[h]
\centering
\scriptsize
\setlength{\tabcolsep}{4pt}
\renewcommand{\arraystretch}{1.0}
\begin{tabular}{p{0.98\linewidth}}
\toprule
\textbf{Task:} Estimate the total number of distinct entities that must be fully known as a complete set to answer this question. Output only an integer.
\par\smallskip
\textbf{Examples:}
\par
Q: Who was the first US president?
\par
A: 1
\par
Q: Which of the 20 amino acids are essential?
\par
A: 20
\par
Q: Which domain is most common across the 88 IAU constellations?
\par
A: 88
\par\smallskip
\textbf{Question:} \{question\_text\}
\par
\textbf{Answer:}
\\
\bottomrule
\end{tabular}
\caption{LLM prompt for knowledge width estimation ($|U|$).}
\label{tab:llm_prompt_width}
\end{table}

\begin{table}[h]
\centering
\scriptsize
\setlength{\tabcolsep}{4pt}
\renewcommand{\arraystretch}{1.0}
\begin{tabular}{p{0.98\linewidth}}
\toprule
\textbf{Task:} Minimize the given DAG by removing redundant nodes and shortening dependency chains.
\par\smallskip
\textbf{Input:}
\par
Question: \{question\}
\par
Nodes: \{nodes\_json\}
\par
Edges: \{edges\_json\}
\par
Final: \{final\_node\}
\par\smallskip
\textbf{Steps:}
\par
1. Remove redundant nodes
\par
2. Merge combinable nodes
\par
3. Shorten chains
\par
4. Maintain correctness
\par\smallskip
\textbf{Output (JSON only):}
\par
\texttt{\{"nodes": [...], "edges": [...], "final\_node": "nK", "optimized": true\}}
\\
\bottomrule
\end{tabular}
\caption{LLM prompt for reasoning depth estimation via operator DAG minimization.}
\label{tab:llm_prompt_dag}
\end{table}

\paragraph{Pooled percentile normalization.}
Let $\Omega$ be the pooled reference set of all benchmark items and let $N=|\Omega|$. For each raw component $x\in\{S_{\mathrm{raw}},W_{\mathrm{raw}},D_{\mathrm{raw}}\}$, we map $x(q)$ to its pooled percentile using mid-rank for ties:
\begin{equation}
\tilde{x}(q)=\frac{\mathrm{rank}_{\Omega}(x(q))-0.5}{N}\in(0,1).
\end{equation}
Applying this transformation separately to the three raw components yields comparable normalized scores $S(q),W(q),D(q)\in(0,1)$.

\paragraph{Iceberg Gap.}
Finally, we define Iceberg Gap as the geometric mean of the three normalized components:
\begin{equation}
\mathrm{IG}(q)=\big(S(q)\cdot W(q)\cdot D(q)\big)^{1/3}\in(0,1).
\end{equation}
This form ensures that IG is high only when surface simplicity, knowledge width, and reasoning depth are all high.

\section{Experimental Setup and Evaluation Protocol}
\label{app:eval_setup}

\subsection{Experimental Setup}
\label{app:exp_setup}

We run two evaluation suites: (i) inference on knowledge-grounded reasoning questions (KRQs) and (ii) enumeration evaluation on enumeration questions (EQs). Both suites use the same model pool and zero-shot prompting, but differ in their scoring protocols. All experiments in this appendix, as well as all subsequent inference experiments, are conducted on $4\times$ NVIDIA H100 94GB GPUs using vLLM.

\begin{table}[h]
\centering
\scriptsize
\setlength{\tabcolsep}{4pt}
\renewcommand{\arraystretch}{1.0}
\begin{tabular}{lcc}
\toprule
\textbf{Setting} & \textbf{EQ} & \textbf{KRQ} \\
\midrule
Scoring & Universe F1 & Accuracy \\
Judge model & Qwen3-30B-A3B-Instruct-2507 & -- \\
\midrule
Decoding & greedy & greedy \\
Temperature & 0.0 & 0.0 \\
Top-$p$ & 1.0 & 1.0 \\
Max tokens & 10{,}240 & 512 \\
Batch size & adaptive & adaptive \\
\bottomrule
\end{tabular}
\caption{Experimental configuration.}
\label{tab:experiment_config}
\end{table}

\subsection{LLM-as-Judge for Enumeration Questions}
\label{app:llm_judge_retrieval}

Enumeration outputs are set-valued, and strict string matching is brittle under aliases, abbreviations, translations, and minor formatting variation. We therefore adopt a hybrid protocol: rule-based matching first, followed by LLM judging only for unresolved cases.

\paragraph{Parsing.}
We parse gold and predicted texts into item lists $\mathcal{G}$ and $\mathcal{P}$ by splitting on common delimiters (commas, semicolons, and newlines, including Chinese variants), stripping list markers (e.g., ``1)'', ``-'', and ``\textbullet''), trimming whitespace, and deduplicating.

\paragraph{Normalization and rule matching.}
We apply lightweight normalization to both gold and predicted items, including Unicode NFKC normalization, whitespace collapsing, lowercasing (when applicable), and trimming trailing punctuation. We then perform exact matching under this normalized form.

\paragraph{LLM judging.}
For predicted items not matched by rules, we invoke an LLM judge to determine whether each predicted item is semantically equivalent to \emph{any} gold item. The judge is instructed to accept aliases, abbreviations, translations, official versus common names, minor spelling variants, and punctuation, case, or diacritic differences, while rejecting loosely related items, different entities with similar names, or partial overlaps that alter meaning. If unsure, the judge must output \texttt{false}. We use Qwen3-30B-A3B-Instruct-2507 with greedy decoding (temperature $=0.0$) and require strictly valid JSON outputs.

\paragraph{One-to-one matched pairs.}
After obtaining judge decisions, we construct a one-to-one matching between $\mathcal{P}$ and $\mathcal{G}$ as follows: (i) rule-matched pairs are fixed; (ii) for the remaining predicted items marked as matchable by the judge, we greedily assign each prediction to at most one gold item, and each gold item to at most one prediction, prioritizing exact or normalized matches when available. Let $m$ denote the resulting number of matched (prediction, gold) pairs.

\paragraph{Universe F1.}
We compute set-level precision, recall, and F1 as
\begin{equation}
\begin{aligned}
\mathrm{Precision} &= \frac{m}{|\mathcal{P}|}, \qquad
\mathrm{Recall} = \frac{m}{|\mathcal{G}|}, \\
\mathrm{F1} &= \frac{2\,\mathrm{Precision}\cdot\mathrm{Recall}}{\mathrm{Precision}+\mathrm{Recall}}.
\end{aligned}
\end{equation}

\subsection{LLM Judge Prompt}
\label{app:judge_prompt}

Table~\ref{tab:llm_prompt_judge} shows the prompt used for the LLM judge. Predicted items are provided as an indexed list, and the judge outputs a boolean decision for each index.

\begin{table}[h]
\centering
\scriptsize
\setlength{\tabcolsep}{4pt}
\renewcommand{\arraystretch}{1.0}
\begin{tabular}{p{0.98\linewidth}}
\toprule
\textbf{Task:} For each predicted item, decide whether it is equivalent to \emph{any} gold item. Accept aliases, abbreviations, translations, minor spelling variants, and punctuation, case, or diacritic differences; reject loosely related or different entities. If unsure, output \texttt{false}.
\par\smallskip
\textbf{Gold items:} \{gold\_items\}
\par
\textbf{Predicted items (indexed):} \{pred\_items\_indexed\}
\par\smallskip
\textbf{Output (JSON only):}
\par
\texttt{\{"pred\_matches":[\{"pred\_index":0,"match":true\},\dots]\}}
\par\smallskip
\textbf{Constraints:} include every \texttt{pred\_index} exactly once; do not add extra keys.
\\
\bottomrule
\end{tabular}
\caption{LLM-as-judge prompt for matching predicted items to a gold list.}
\label{tab:llm_prompt_judge}
\end{table}

\section{Error Study of Gemini-3-Flash}
\label{app:error_study}

This appendix reports CoT results for two closed-source models and characterizes residual error types through a manual analysis.

\subsection{Closed-Source CoT Evaluation}
\label{app:closed_source_cot}

\paragraph{Models and data.}
We evaluate DeepSeek-Chat and Gemini-3-Flash on English KRQs from \textsc{KnowledgeBerg} under the same CoT setting used in \S\ref{subsec:diagnosis}.

\paragraph{Results.}
DeepSeek-Chat achieves 48.39\% accuracy, and Gemini-3-Flash achieves 65.24\% accuracy.

\subsection{Manual Error Taxonomy for Gemini-3-Flash}
\label{app:gemini_error_taxonomy}

\paragraph{Sample.}
To characterize residual failures in the strongest tested setting, we randomly sample $N=100$ incorrect predictions from Gemini-3-Flash.

\paragraph{Labels.}
Each error is assigned to exactly one category:
\begin{itemize}[leftmargin=*,noitemsep,topsep=2pt]
    \item \textbf{Completeness:} the model lacks, omits, or states incorrect universe elements required for the question.
    \item \textbf{Awareness:} the model fails to recognize the required knowledge and instead focuses narrowly on the options or surface cues.
    \item \textbf{Application:} the model identifies relevant knowledge but applies an incorrect set operation or composition (e.g., wrong filtering, miscounting, or invalid aggregation/comparison).
    \item \textbf{Artifact:} parsing errors, ill-formed answers, or other output-format issues.
\end{itemize}

\paragraph{Category proportions.}
Across 100 randomly sampled Gemini-3-Flash CoT errors, the distribution is as follows: \textit{Completeness} 42\%, \textit{Awareness} 17\%, \textit{Application} 38\%, and \textit{Artifact} 3\%.

\section{Testing-Time Compute: Configurations}
\label{app:testing_time_compute}

This appendix documents the inference-time compute strategies used in our experiments: \textit{Self-Consistency}, \textit{Self-Refine}, \textit{Self-Verification}, and \textit{Retrieval-Augmented Generation} (RAG). We report their key hyperparameters, decoding settings, token budgets, aggregation or selection rules, and prompt templates.

\subsection{Method Configurations}
\label{app:ttc_configs}

Table~\ref{tab:ttc_config} summarizes the configurations used in all testing-time compute experiments.

\begin{table}[h]
\centering
\scriptsize
\setlength{\tabcolsep}{3pt}
\renewcommand{\arraystretch}{1.05}
\begin{tabularx}{\columnwidth}{l p{0.34\columnwidth} >{\raggedright\arraybackslash}X}
\toprule
\textbf{Method} & \textbf{Key hyperparameters} & \textbf{Decoding / budget} \\
\midrule
Self-Consistency &
$N\in\{4,8,16\}$;\newline
majority vote &
\textit{Gen:} $T{=}0.8$, top-$p{=}0.95$\newline
max tokens: \newline 4096 ($N{=}4$);\newline
2048 ($N{\in}\{8,16\}$) \\
\addlinespace[2pt]
\hline
Self-Refine &
$K\in\{1,2,3\}$ rounds;\newline
Solve $\rightarrow$ Critique $\rightarrow$ Revise &
\textit{Solve/Revise:} $T{=}0.8$, top-$p{=}0.95$\newline
max tokens: 4096\newline
\textit{Critique:} $T{=}0.0$, top-$p{=}1.0$\newline
max tokens: 512 \\
\addlinespace[2pt]
\hline
Self-Verification &
Best-of-$N$, \newline $N\in\{4,8,16\}$;\newline
verifier selects $\arg\max$ &
\textit{Gen:} $T{=}0.8$, top-$p{=}0.95$\newline
max tokens: 4096\newline
\textit{Verify:} $T{=}0.0$, top-$p{=}1.0$\newline
max tokens: 64 \\
\addlinespace[2pt]
\hline
RAG &
Retriever: \newline dense / hybrid / rerank;\newline
$k{=}8$ retrieved passages &
\textit{Gen:} $T{=}0.1$, top-$p{=}1.0$\newline
max tokens: 4096\newline
retrieved context $\leq$ 12k chars \\
\bottomrule
\end{tabularx}
\caption{Testing-time compute configurations. ``Gen'' denotes answer generation; auxiliary steps use deterministic decoding unless otherwise noted. ``max tokens'' denotes the maximum generation length per sample or step.}
\label{tab:ttc_config}
\end{table}

\begin{table}[t]
\centering
\scriptsize
\setlength{\tabcolsep}{3pt}
\renewcommand{\arraystretch}{0.95}
\begin{tabular}{p{0.22\linewidth} p{0.74\linewidth}}
\toprule
\textbf{Prompt} & \textbf{Template} \\
\midrule
\textbf{Shared generation prompt} &
\textbf{Task:} Answer the following multiple-choice question by selecting the correct option.\newline
Question: \{question\}\newline
Options:\newline
\{options\}\newline
\textbf{Requirement:} Think step by step, then output \texttt{\textbackslash boxed\{X\}}, where $X$ is a single option label.
\\
\midrule
\textbf{Self-Refine:} \emph{Critique} &
\textbf{Task:} Critique the solution attempt. Identify logical mistakes, missing assumptions, or incorrect option comparisons. Provide actionable feedback.\newline
Question: \{question\}\newline
Options:\newline
\{options\}\newline
Solution attempt: \{attempt\}
\\
\midrule
\textbf{Self-Refine:} \emph{Revise} &
\textbf{Task:} Revise the solution to address the critique. Then output \texttt{\textbackslash boxed\{X\}}.\newline
Question: \{question\}\newline
Options:\newline
\{options\}\newline
Previous attempt: \{attempt\}\newline
Critique: \{critique\}
\\
\midrule
\textbf{Self-Verification:} \emph{Verifier} &
\textbf{Task:} Select the single best candidate. Output only \texttt{\textbackslash boxed\{k\}}, where $k\in\{1,\dots,n\}$.\newline
Question: \{question\}\newline
Options:\newline
\{options\}\newline
Candidates:\newline
\{candidates\_block\}
\\
\midrule
\textbf{RAG} &
\textbf{Retrieved context (may be partial or irrelevant):}\newline
\{context\}\newline
\textbf{Task:} Answer the following multiple-choice question by selecting the correct option.\newline
Question: \{question\}\newline
Options:\newline
\{options\}\newline
\textbf{Requirement:} Think step by step, then output \texttt{\textbackslash boxed\{X\}}.
\\
\bottomrule
\end{tabular}
\caption{Prompt templates for testing-time compute methods.}
\label{tab:ttc_prompts}
\end{table}

\paragraph{Self-Consistency.}
For each question $q$, we sample $N\in\{4,8,16\}$ independent reasoning trajectories under stochastic decoding and aggregate the final answers by majority vote. To control compute at larger $N$, we cap generation length at 4096 tokens for $N=4$ and 2048 tokens for $N\in\{8,16\}$. Ties are resolved by selecting the earliest generated candidate.

\paragraph{Self-Refine.}
We implement iterative refinement as a fixed loop:
\[
\textit{Solve} \rightarrow (\textit{Critique} \rightarrow \textit{Revise}) \times K,
\]
with $K\in\{1,2,3\}$. Solve and Revise use stochastic decoding ($T=0.8$, top-$p=0.95$) with a 4096-token budget, while Critique uses deterministic decoding ($T=0.0$, top-$p=1.0$) with a 512-token budget.

\paragraph{Self-Verification.}
We decouple proposal from selection. The model first generates $N\in\{4,8,16\}$ candidate solutions under $T=0.8$ and top-$p=0.95$, with a 4096-token budget. A verifier then selects the best candidate under deterministic decoding ($T=0.0$, top-$p=1.0$), producing a compact 64-token output that specifies only the selected candidate index.

\paragraph{RAG.}
RAG prepends retrieved evidence to the prompt at inference time. Dense retrieval uses sentence-transformer embeddings (\texttt{all-MiniLM-L6-v2}) and retrieves $k=8$ passages per question, prepending up to 12{,}000 characters of retrieved context. Hybrid retrieval combines BM25 and dense scores with equal weighting ($\alpha=0.5$). For reranking, we first retrieve $k=50$ candidates and then apply a cross-encoder reranker (\texttt{cross-encoder/ms-marco-MiniLM-L-6-v2}) before selecting the top 8. Documents are chunked into 800-token segments with 100-token overlap. Generation uses low temperature ($T=0.1$, top-$p=1.0$) to encourage evidence-grounded outputs.

\end{document}